\documentclass{article}
\usepackage{PRIMEarxiv}
\usepackage[utf8]{inputenc} 

\usepackage{amsmath,amsfonts}
\usepackage{algorithmic}
\usepackage{algorithm}
\usepackage{array}
\usepackage[caption=false,font=normalsize,labelfont=sf,textfont=sf]{subfig}
\usepackage{textcomp}
\usepackage{stfloats}
\usepackage{url}
\usepackage{verbatim}
\usepackage{graphicx}
\usepackage{booktabs}
\usepackage[flushleft]{threeparttable}
\usepackage{multirow}
\usepackage{setspace}
\usepackage{makecell} 
\usepackage{soul, color}
\usepackage{flushend}
\usepackage{ulem}
\usepackage{cite}
\usepackage{enumitem}
\usepackage{hyperref}
\hypersetup{hidelinks}
\hypersetup{
        colorlinks=true,
        linkcolor=blue,
        filecolor=green,      
        urlcolor=blue,
        citecolor=red,
}

\title{Interpreting What Typical Fault Signals Look Like via Prototype-matching}

\author{
  Qian Chen \\
  Shanghai Jiao Tong University \\
  \texttt{chenqian2020@sjtu.edu.cn} \\
     \And
  Xingjian Dong \thanks{Corresponding Author.\\ This work was supported by the National Natural Science Foundation of China under Grant No. 12272219, and No. 12121002.} \\
  Shanghai Jiao Tong University \\
  \texttt{donxij@sjtu.edu.cn} \\
     \And
  Zhike Peng\\
  Shanghai Jiao Tong University \\
  \texttt{z.peng@sjtu.edu.cn} \\
}

\begin{document}
\maketitle

\begin{abstract}
        Neural networks, with powerful nonlinear mapping and classification capabilities, are widely applied in mechanical fault diagnosis to ensure safety. However, being typical black-box models, their application is limited in high-reliability-required scenarios. To understand the classification logic and explain what typical fault signals look like, the prototype matching network (PMN) is proposed by combining the human-inherent prototype-matching with autoencoder (AE). The PMN matches AE-extracted feature with each prototype and selects the most similar prototype as the prediction result. It has three interpreting paths on classification logic, fault prototypes, and matching contributions. Conventional diagnosis and domain generalization experiments demonstrate its competitive diagnostic performance and distinguished advantages in representation learning. Besides, the learned typical fault signals (i.e., sample-level prototypes) showcase the ability for denoising and extracting subtle key features that experts find challenging to capture. This ability broadens human understanding and provides a promising solution from interpretability research to AI-for-Science.
\end{abstract}


\section{Introduction}

Mechanical fault diagnosis is pervasive in both industrial production and equipment maintenance, aiming to mitigate property damage and improve production efficiency.
The development of sensors and IoT devices has led to the generation of a massive amount of operational data, marking the advent of big data era~\cite{leiApplicationsMachineLearning2020}.

Traditional fault diagnosis, which relies on expert knowledge and physical models, necessitates extensive prior knowledge and involves manual feature construction and selection. So it is unsuitable for the complex and unpredictable scenarios presented by big data.
In contrast, leveraging this abundant data, intelligent fault diagnosis (IFD), adept at big data processing and independent of prior knowledge, has gained prominence in the field of mechanical fault diagnosis~\cite{zhaoDeepLearningAlgorithms2020}.

IFD, harnessing the nonlinear mapping capabilities of neural networks, offers an end-to-end solution for fault diagnosis. Various advanced networks have been sequentially applied in the fault diagnosis of mechanical equipment, including gears, bearings, and more. Despite the unparalleled advantages of IFD in diagnostic accuracy~\cite{michauFullyLearnableDeep2022}, information fusion~\cite{liFusionCWSMMBasedFramework2022} and other aspects, it still exhibits a significant drawback—poor interpretability~\cite{machlevMeasuringExplainabilityTrustworthiness2022}.

Neural network, serving as the core of IFD, is a typical \textit{black-box} that lacks interpretability. Key aspects of which, such as the decision basis, classification logic, and the meaning of weights remain unknown. Moreover, the interpretability is the crucial foundation for the practical application of IFD. From a user perspective, uninterpretable IFD demands additional solutions for cross-validation, leading to increased costs. From a developer standpoint, uninterpretable IFD poses more challenges for correcting diagnostic errors and enhancing model performance. From the view of potential risks, real-world fault diagnosis scenarios are far more complex and unpredictable than laboratory condition, making it difficult for uninterpretable models to guarantee their performance in practical work environments. Therefore, enhancing model interpretability becomes a key challenge for IFD.

Interpretability research has flourished in computer vision (CV)~\cite{ivanovsPerturbationbasedMethodsExplaining2021} and natural language processing (NLP)~\cite{madsenPosthocInterpretabilityNeural2023}, but it is still in its early stages in the field of IFD. Current interpretability research of IFD primarily focuses on the attribution aspect, which means \textit{assigning credit (or blame) to the input features (e.g., feature importance, saliency masks)}~\cite{zhangSurveyNeuralNetwork2021}. Some scholars, inspired by interpretability researches in CV, have extended them to the IFD. For instance, Wu et al.~\cite{wuHybridClassificationAutoencoder2021} and Grezmak et al.~\cite{grezmakInterpretableConvolutionalNeural2020} utilized existing class activation map (CAM) and layer-wise relevance propagation (LRP) attribution methods to achieve time-frequency domain attribution for diagnosis models of bearing and motor. However, these methods are designed for image data, potentially resulting in suboptimal interpretability when applied to vibration signals. Furthermore, some scholars have taken a step further by leveraging the characteristics of vibration signals to construct more targeted interpreting methods for IFD. Specifically,
our previous work~\cite{qianchenTFNInterpretableNeural2024}
embeded time-frequency transform (TFT) into the traditional convolutional layer, thereby utilizing amplitude-frequency response to unveil the network's attention to different frequencies. Tang et al.~\cite{tangSignalTransformerRobustInterpretable2022} and Li et al.~\cite{liVariationalAttentionBasedInterpretable2022} employed the multi-head attention mechanism of the Transformer architecture to reveal the model's attention to different segments of the temporal sample. Li et al.~\cite{liMultilayerGradCAMEffective2023} enhanced traditional CAM through correlation weighting, giving emphasis to fault characteristic frequencies in the CAM results.

However, the diagnostic logic has been largely neglected by current interpretability research in IFD, with limited related work in this area. An et al.~\cite{anAdversarialAlgorithmUnrolling2023} utilized algorithm unrolling to learn a dictionary set of fault components (i.e., rotational frequency, meshing frequency) from the network perspective, but these learned components are at a local level rather than representing the complete fault signals.

\begin{figure}[htbp]
        \centering
        \includegraphics[width=10 cm]{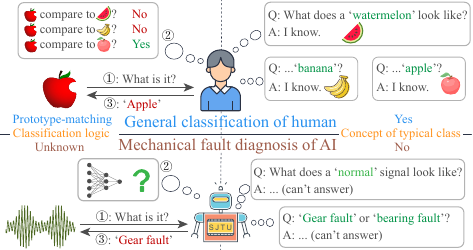}
        \caption{The classification logic of prototype-matching,  a concept that categorizes an object based on its
                similarity to the prototype. Prototype-matching is a common logic in general human classification, where human can tell the typical sample of each class. In contrast, the classification logic of AI in IFD tasks remains unknown, despite its excellent diagnostic prediction, and AI could not explain ''\textit{what typical fault signals look like}''.}
        \label{fig:intro}
\end{figure}

To understand the classification logic and represent complete fault signals, we narrow our focus to the prototype-matching, a concept that categorizes an object based on its similarity to the prototype. As depicted in Fig.~\ref{fig:intro}, prototype-matching is inherently rooted in human classification logic, but neural networks (or AI) do not possess this capability naturally.

The application of prototype-matching could be roughly devided as three stages. In the initial stage, prototype-matching is applied to original samples or maually fabricated features (e.g., \textit{K-means} or \textit{mixture density estimation}), but these methods lack the nonlinear mapping ability and are only suitable for simple tasks. In the second stage, prototype-matching is applied to the embedded features of neural networks (i.e., \textit{Prototypical Network}~\cite{snellPrototypicalNetworksFewshot2017}), allowing it to handle more complex tasks. Nonetheless, its prototypes are simply the mean of embedded features, and the interpretability of feature-level prototype is quite limited. In the current stage, Li et al. treat the prototype as a learnable vector and use the decoded prototype to interpret the class prototype~\cite{liDeepLearningCaseBased2018} and semantic prototype~\cite{chenThisLooksThat2019} of images. However, the above methods are mainly focused on the CV area and cannot be directly applied to IFD domain due to the significant differences between visual images and vibration signals in terms of interpretive form and acceptability.

In recent years, prototype-matching, proven to br effective for few-shot learning by \cite{snellPrototypicalNetworksFewshot2017}, has been introduced into IFD to address the problem of limited samples~\cite{liMultiscaleDynamicFusion2020,jiangFewshotFaultDiagnosis2022,sunAdaptiveAntinoiseGear2022}. With its excellent performance, prototype-matching has been applied to a wider range of complex problems, including semi-supervised learning~\cite{zhangSemisupervisedMomentumPrototype2022,zhouSemiSupervisedFederatedLearning2023,suSemisupervisedWeightedCentroid2024}, transfer learning~\cite{wangDeepPrototypicalNetworks2020,yangNovelBrownianCorrelation2022,tangImprovedPrototypicalNetwork2023,chenNovelMomentumPrototypical2023,zhangDualPrototypicalContrastive2023,sunOpenSetDiagnosis2023,longMultidomainVariancelearnablePrototypical2023,meiCrossdomainOpensetFault2024}, federal learning~\cite{wangFederatedContrastivePrototype2023,zhouSemiSupervisedFederatedLearning2023} and other aspects.
We conducted a detailed literature review on the application of prototype-matching in IFD as listed in Table~\ref{tab:literature}.
However, although numerous research about prototype-matching have been conducted in many aspects, the huge potential of prototype-matching in interpretability is still overlooked. Therefore, exploring the interpretability of prototype-matching in IFD domain is a topic that remains to be explored and is highly innovative.

\begin{table}[htbp]
        \centering
        \footnotesize
        \begin{threeparttable}
                \caption{The literature that apply prototype-matching in IFD domain. \label{tab:literature}}
                \setlength\tabcolsep{14pt} 
                \begin{tabular*}{\hsize}{@{\extracolsep{\fill}}ll}
                        \toprule[1pt]
                        \multicolumn{1}{c}{\textbf{Focusing aspect of prototype-matching}} & \multicolumn{1}{l}{\textbf{Literature}} \\
                        \midrule[0.3pt]
                        Few   shot                                  & \cite{liMultiscaleDynamicFusion2020,jiangFewshotFaultDiagnosis2022,sunAdaptiveAntinoiseGear2022}       \\
                        Few   shot + Semi-supervised learning              & \cite{zhangSemisupervisedMomentumPrototype2022,zhouSemiSupervisedFederatedLearning2023,suSemisupervisedWeightedCentroid2024}          \\
                        Few   shot   +   Close-set transfer learning & \cite{wangDeepPrototypicalNetworks2020,yangNovelBrownianCorrelation2022,tangImprovedPrototypicalNetwork2023,chenNovelMomentumPrototypical2023,zhangDualPrototypicalContrastive2023}          \\
                        Few   shot   +   Open-set transfer learning  & \cite{sunOpenSetDiagnosis2023,longMultidomainVariancelearnablePrototypical2023,meiCrossdomainOpensetFault2024}          \\
                        Few   shot   +   Federal learning          & \cite{wangFederatedContrastivePrototype2023,zhouSemiSupervisedFederatedLearning2023}             \\
                        Few   shot + Prototype drifting suppression & \cite{zhangSemisupervisedMomentumPrototype2022,chenNovelMomentumPrototypical2023,suSemisupervisedWeightedCentroid2024}          \\
                        Other   aspects (Meta-learing, Anti-noise)   & \cite{sunAdaptiveAntinoiseGear2022,tangImprovedPrototypicalNetwork2023}            \\
                        Interpretability                            & Our work only\\
                        \bottomrule[1pt]
                \end{tabular*}
        \end{threeparttable}
\end{table}

Given the aforementioned issue, we introduce prototype-matching into the IFD domain and propose prototype-matching network (PMN) with its comprehensive application framework. Essentially, the proposed PMN model is a combination of autoencoder (AE) and a classifier based on the prototype-matching layer (PM-Layer). The dimensionality reduction of AE provides feasibility for prototype-matching, and the aggregative nature of prototype-matching augments the representation learning capabilities of AE.


As for interpretability, the proposed PMN has three meaningful interpreting paths. \textbf{1) Classification logic}: PMN is based on the intuitionistic prototype-matching and could naturally explain ``\textit{what the classification logic is}''. \textbf{2) Sample-level prototype}: PMN learns the prototype in feature-level and reconstructs it into sample domain by the decoder, which explains \textit{``what typical fault signals look like}''. \textbf{3) Similarity source}: the contribution of each frequency is measured, that explains ``\textit{which frequency is responsible for the similarity of input vibration sample with the matched prototype}''.

The main contributions could be summarized as follows:
\begin{enumerate}
        \item Prototype-matching is combined with autoencoder to formulate the prototype-matching network (PMN), and is firstly applied to improve the interpretability of IFD.
        \item The proposed PMN has three interpreting paths: the classification logic, the typical fault signals, and the frequency contributions to the matching similarity.
        \item Conventional diagnosis and domain generalization experiments are conducted to validate the performance of the proposed PMN in terms of diagnostic accuracy, representation learning, and interpretability.
\end{enumerate}

\section{Methodology}

\subsection{Model Achitecture}

\begin{figure}[htbp]
        \centering
        \includegraphics[width=10 cm]{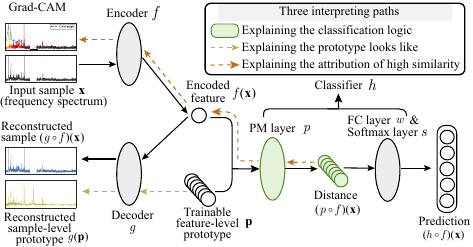}
        \caption{The Architecture of Proposed PMN.}
        \label{fig:Method}
\end{figure}

Let $D^{train}=\{({\mathbf x}_i,y_i)\}_{i=1}^{n}$ denotes the training dataset with sample ${\mathbf x} \in \mathbb{R}^p$ and label $y \in \{1,2,\cdots,K\}$. As shown in Fig.~\ref{fig:Method}, the proposed PMN consists of three components, the encoder network $f:\mathbb{R}^p \rightarrow \mathbb{R}^q$, the decoder network $g:\mathbb{R}^q \rightarrow \mathbb{R}^p$, and the classifier network $h:\mathbb{R}^q \rightarrow \mathbb{R}^K$. Like the traditional autoencoder, the encoder $f$ takes input signal sample $\mathbf x$ (i.e., the frequency spectrum) into encoded feature $f(\mathbf{x}) \in \mathbb{R}^q$ and the decoder $g$ recovers it into reconstructed sample $(g\circ f)(\mathbf{x})$. Based on the encoding and decoding process, the encoded feature $f(\mathbf{x})$ could obtain key information from the input frequency spectrum $\mathbf x$ in a low-dimentional latent space. Thus, the fault classification is conducted by the classifier $h$ based on the the encoded feature $f(\mathbf{x})$.

The above structure is quite common in other works~\cite{wuHybridClassificationAutoencoder2021,shaoModifiedStackedAutoencoder2022}, while our innovation lies in the classifier. The classifier consists of a prototype-matching layer (PM-Layer), a full-connected (FC) layer and a softmax layer. The PM-Layer initializes $m$ vectors $\mathbf{p} \in \mathbb{R}^q$ in the latent space as trainable feature-level prototypes. The PM-Layer measures the distance between the encoded feature $\mathbf{z}=f(\mathbf{x}_i)$ and each of the feature-level prototypes:
\begin{equation}
        p(\mathbf{z}) = \left[ d(\mathbf{z},\mathbf{p}_1), d(\mathbf{z},\mathbf{p}_2),\cdots,d(\mathbf{z},\mathbf{p}_m) \right]^T \in \mathbb{R}^m
        \label{eq:Achitecture_PML}
\end{equation}
where, $d(\cdot,\cdot)$ denotes the distance metric. We recommand using the square Euclidean distance $d_{L_2}(x,y)= \left \| x-y \right \|^2_2$ as the distance metric, because it performs slightly better than ${Cosine}$ distance and ${L_1}$ distance according to peer literature~\cite{snellPrototypicalNetworksFewshot2017} and subsection~\ref{subsec:metric}.

The FC layer maps the distance result $p(\mathbf{z}) $ into $K$-dims class space as the predict logits by
\begin{equation}
        \mathbf{v}=W p(\mathbf{z}) \in \mathbb{R}^K, \, {\mathrm where} \  W \in \mathbb{R}^{K\times m}.
        \label{eq:Achitecture_FC}
\end{equation}
To connect eack prototype with a clear fault class (i.e., each column of $W$ is close to a one-hot vector), and considering the negative correlation between the distance and the class probability,
we initialize $W$ by $W_{i,j}= -\mathbb{I} \left( {\rm mod}(j,K)=i \right)$ where the initialized $W$ is equal to a negative identity matrix $-I$ when $m=K$.

The softmax layer $s$ normalizes the logit result $\mathbf{v}$ into a probability distribution over the $K$ fault classes. The probability of $k$-th fault class could be calculated as
\begin{equation}
        s(\mathbf{v})_k = \frac{\exp{(\mathbf{v}_k)}}{\sum_{k'=1}^{K}{\exp{(\mathbf{v}_{k'})}}}
        \label{eq:Achitecture_softmax}
\end{equation}

Essentially, this classifier is distance-based in the low-dimentional feature-level space like \textit{mixture density estimation} (MDE) algorithm, predicting the class based on the nearest prototype to the input. The number of prototypes $m$ is equivalent to the number of density components in MDE. Each fault class has one prototype when $m=K$, or multiple prototypes when $m>K$. The setting of $m$ is a hyperparameter of PMN and is recommanded to set $m=K$ due to the enough nonlinear mapping ability of AE, which is discussed in subsection~\ref{subsec:protonumber}.

\subsection{Loss Function}

The training objectives include both accuracy and interpretability. The former is achieved through traditional classification loss $L_{\mathrm{cla}}$ and the reconstruction loss $L_{\mathrm{recon}}$ of the autoencoder. The latter, encouraging each sample could find a sufficiently close prototype, is realized by three new regularization terms, i.e., $R_{1}$, $R_{2}$ and $R_{3}$.

The classification loss $L_{\mathrm{cla}}$ employs standard cross-entropy to penalize misclassification whose expression is given by
\begin{equation}
        L_{\mathrm{cla}} (h\!\circ \!f, D) = - \frac{1}{n} \sum_{i=1}^n \sum_{k=1}^{K} {\mathbb{I}(y_i\!=\!k) \cdot \log{ \left[
                        (h\!\circ \!f)(\mathbf{x}_i)
                        \right]_k}}.
        \label{eq:L_cla}
\end{equation}

The reconstruction loss $L_{\mathrm{recon}}$ utilizes mean squared error to ensure the reconstructed sample remains consistent with the input sample. This can be denoted as
\begin{equation}
        L_{\mathrm{recon}} (g \circ f, D) = \frac{1}{n}  \sum_{i=1}^n \left \| \mathbf{x}_i - (g \circ f) (\mathbf{x}_i) \right \|^2_2.
        \label{eq:L_recon}
\end{equation}

Regarding interpretability, $R_1$ and $R_2$ minimize the mutual minimum distance between encoded features and prototypes in both views, respectively. They are calculated by
\begin{equation}
        R_{1} (\mathbf{p}, f, D) = \frac{1}{n} \sum_{i}^n \min_{j\in[1,m]} d \left( f(\mathbf{x}_i),\mathbf{p}_j \right)
        \label{eq:R_1}
\end{equation}
\begin{equation}
        R_{2} (\mathbf{p}, f, D) = \frac{1}{m} \sum_{j}^m \min_{i\in[1,n]} d \left( f(\mathbf{x}_i),\mathbf{p}_j \right).
        \label{eq:R_2}
\end{equation}

$R_1$ promotes encoded features to closely approximate a specific prototype, causing the features to cluster around the prototypes. $R_2$ encourages prototypes to closely align with at least one sample in the feature space, facilitating a meaningful reconstruction of prototypes after processed by the decoder. Notably, $R_2$ necessitates global minimization over the entire dataset, which is impractical for large datasets. Hence, we relax this minimization to the local minibatch for simplification.

To ensure the discrimination of prototypes, $R_3$ maximizes the mutual minimum distance between prototypes by
\begin{equation}
        R_{3} (\mathbf{p}) = - \frac{1}{m} \sum_{i}^m \min_{j\in[1,m]} d \left( \mathbf{p}_i,\mathbf{p}_j \right).
        \label{eq:R_3}
\end{equation}

Putting the above five losses together, the total loss $L$ could be expressed as
\begin{equation}
        \begin{split}
                L(g\! \circ \! h \! &\circ\! f,\! \mathbf{p},\! D) = L_{\mathrm{cla}} (h \! \circ \! f, D) + \lambda L_{\mathrm{recon}} (g \!\circ \!f, D) \\
                &+ \lambda_1 R_{1} (\mathbf{p},\! f,\! D) + \lambda_2 R_{2} (\mathbf{p},\! f,\! D) + \lambda_3 R_{3} (\mathbf{p}).
        \end{split}
        \label{eq:L}
\end{equation}
where $\lambda$, $\lambda_1$, $\lambda_2$, $\lambda_3$ are hyperparameters to balance the ratio of different losses.

\subsection{Three Interpreting Paths and Diagnosis Framework}
As shwon in Fig.~\ref{fig:Method}, the first interpreting path is the clear classification logic within the physical-meaningful PM-layer. The PM-layer $p$ compares encoded feature $f(\mathbf{x})$ with each feature-level prototype $\mathbf{p}$ and adopts the fault class of the most similar prototype as the prediction. This classification follows the inherent prototype-matching logic of humans, leading the \textit{black-box} neural network into a \textit{gray-box}.

The second interpreting path is interpreting ``\textit{what typical fault signals look like}". Due to the synergistic combination of autoencoder and PM-Layer, feature-level prototypes $\mathbf{p}$ are learned by the PM-Layer and reconstructed into the sample domain by the decoder. The  outcome $g(\mathbf{p})$, sample-level prototypes, explains the typical fault signal looks like from the model's perspective.

The third interpreting path is explaining the attribution of high similarity through attribution-type interpreting methods. these methods assign credit to each node of the input frequency spectrum and  Grad-CAM \cite{selvarajuGradCAMVisualExplanations2017} is employed in our work. The result reveals the crucial fault-related frequency that contributes to the high similarity between the input signal sample and the matched prototype.

\begin{figure*}[t!]
        \centering
        \includegraphics[width=17 cm]{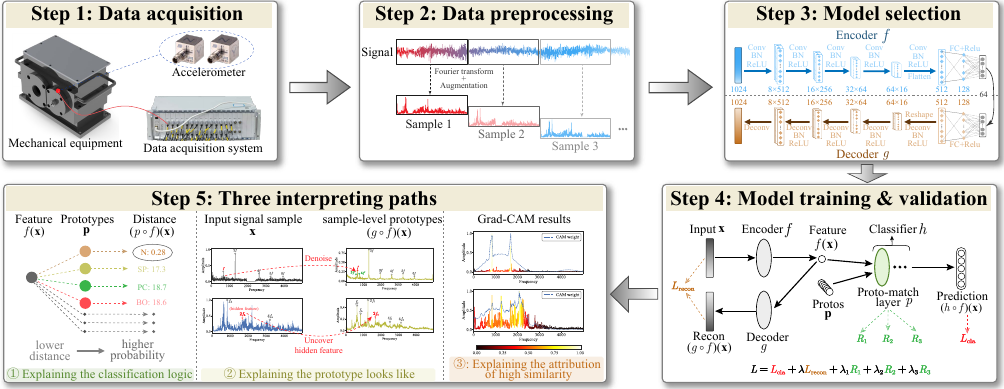}
        \caption{The entire framework for applying PMN to mechanical intelligent fault diagnosis.}
        \label{fig:Framework}
\end{figure*}

The diagnosis framwork of PMN is illustrated in Fig.~\ref{fig:Framework}. Firstly, the mechanical vibration signals are collected by accelerometers, trancated and transformed into frequency domain using Fourier transform as training and test samples. Secondly, the autoencoder structure can be selected as the backbone for the classifier, and we choose a relatively shallow CNN-based autoencoder 
 as shown in Table.~\ref{tab:autoencoder structure}.
Subsequently, the PMN is trained by Eq.~(\ref{eq:L}) on the training dataset $\mathcal{D}^\textrm{train}$, and the performance of PMN is evaluated using the test dataset $\mathcal{D}^\textrm{test}$. Finally, three interpreting paths are conducted on the trained PMN to explain the classification logic, depict the typical fault signals and pinpoint the crucial fault-related frequency causing high similarity with matched prototype in model's view.

\begin{table}[htbp]
        \centering
        \begin{threeparttable}
                \caption{The Model Architecture Used in Experiments.\label{tab:autoencoder structure}}
                \small

                \begin{tabular*}{\hsize}{@{\extracolsep{\fill}}lllll}
                        \toprule[1pt]
                        Part & No. & Basic Unit                                    & Output size     \\
                        \midrule[0.3pt]
                        \vspace{1mm}
                        Enc. & -   & Input                                         & 1 $\times$ 1024 \\
                             & 1   & Conv(9@2@4)\tnote{\dag} -BN-ReLU              & 8 $\times$ 512  \\
                             & 2   & Conv(9@2@4)-BN-ReLU                           & 16 $\times$ 256 \\
                             & 3   & Conv(11@4@5)-BN-ReLU                          & 32 $\times$ 64  \\
                             & 4   & Conv(11@4@5)-BN-ReLU                          & 64 $\times$ 16  \\
                             & 5   & Conv(11@4@5)-BN-ReLU-Flatten                  & 128 $\times$ 4  \\
                             & 6   & FC(128)-ReLU-FC(64)                           & 64              \\
                        \midrule[0.3pt]
                        Dec. & 1   & FC(128)-ReLU-FC(512)-Reshape                  & 128 $\times$ 4  \\
                             & 2   & Deconv(10@4@3)-BN-ReLU                        & 64 $\times$ 16  \\
                             & 3   & Deconv(10@4@3)-BN-ReLU                        & 32 $\times$ 64  \\
                             & 4   & Deconv(8@2@3)-BN-ReLU                         & 16 $\times$ 256 \\
                             & 5   & Deconv(8@2@3)-BN-ReLU                         & 8 $\times$ 512  \\
                             & 6   & Deconv(8@2@3)                                 & 1024            \\
                        \midrule[0.3pt]
                        Cla. & 1   & PMLayer($m$\tnote{\ddag} )-FC($K$\tnote{\S} ) & $K$             \\
                        \bottomrule[1pt]
                \end{tabular*}

                \begin{tablenotes}
                        \smallskip
                        \footnotesize
                        \item[\dag] \textit{(x@y@z)}: \textit{x}, \textit{y}, and \textit{z} represent the kernel size, stride, and padding, respectively.
                        \item[\ddag] $m$: the number of prototypes as a hyperparameter of PM layer.
                        \item[\S] $K$: the number of fault classes and determined by the dataset.
                \end{tablenotes}
        \end{threeparttable}
\end{table}

\section{Experiments}

\subsection{Conventional Fault Diagnosis of Planetary Gearbox}

\begin{figure}[htbp]
        \centering
        \includegraphics[width=10 cm]{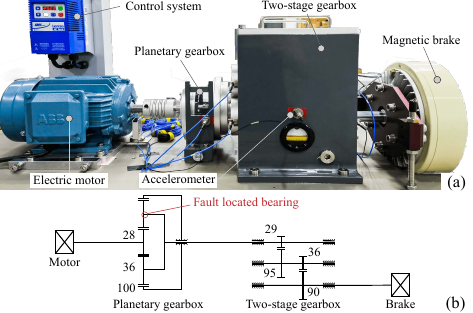}
        \caption{The experimental setup for the planetary gearbox dataset. a) The Experiment rig. b) The schematic diagram.}
        \label{fig:Exp1-rig}
\end{figure}

{\bfseries \itshape 1) Data preparation}: The experimental rig for the planetary gearbox dataset is depicted in Fig.~\ref{fig:Exp1-rig}, comprising an electric motor, planetary gearbox, two-stage gearbox, magnetic brake, accelerometer, and other components. The accelerometer is installed on the shell of the planetary gearbox to gather vibration signals at a sampling frequency of 12 kHz. The speed of the electric motor is set to 1800 rpm.

\begin{figure}[htbp]
        \centering
        \includegraphics[width=10 cm]{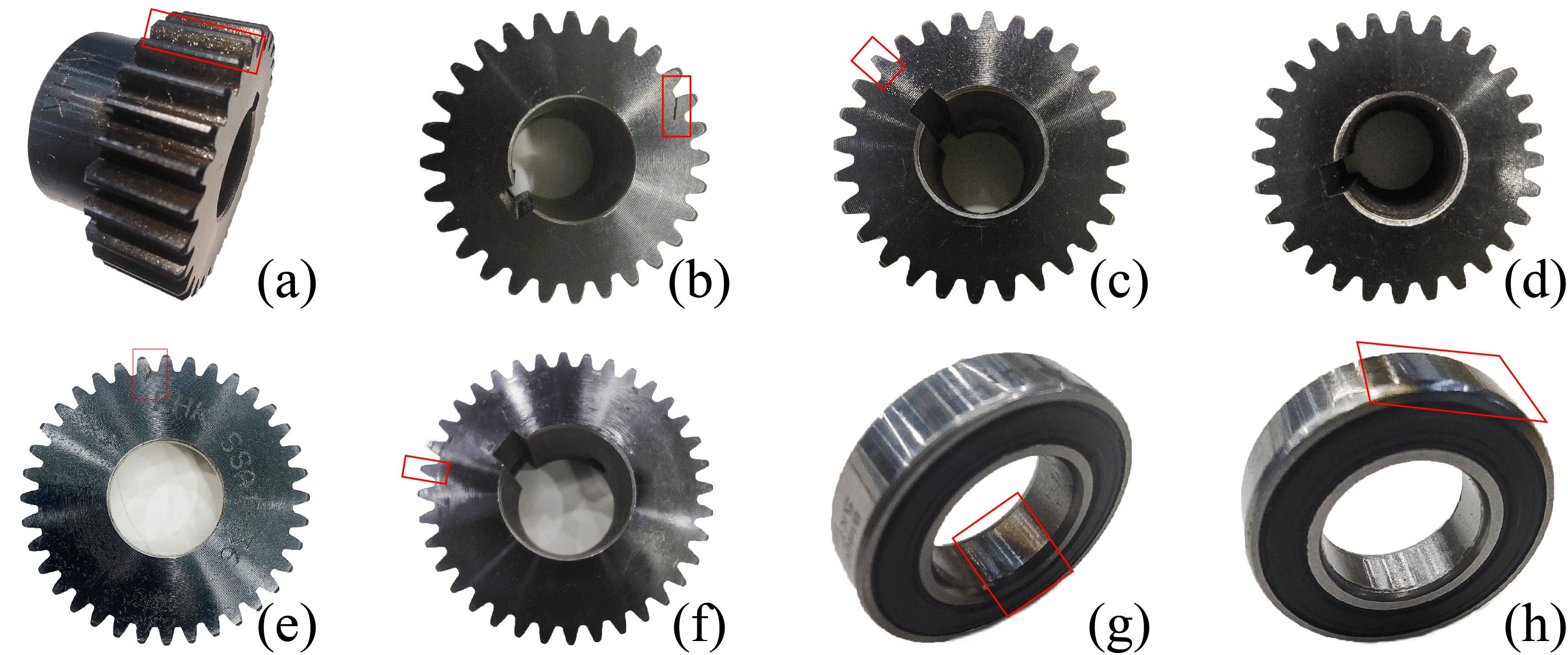}
        \caption{Ten fault types included in the planetary gearbox dataset. a) SP: Sun gear pitting. b) SC: Sun gear crack. c) Sw: Sun gear partial tooth wearing. d) SW: Sun gear full tooth wearing. e) PC: Planet gear crack. f) PW: Planet gear full tooth wearing. g) BI: Bearing inner race. h) BO: Bearing outer race. Additionally, two other faults, bearing cage (BC) fault rolling (BR) fault, are also considered.}
        \label{fig:Exp1-fault}
\end{figure}

The dataset encompasses ten types of mechanical faults as illustrated in Fig.~\ref{fig:Exp1-fault}, including six types of planetary gearbox faults (SP, SC, Sw, SW, PC, PW) and four types of bearing faults (BI, BO, BC, BR). considering the normal state, this experiment could be regarded as a 11-class classification task.

During data preparation, the raw vibration signal is trancated without overlap using a sliding window  and is subsequently transformed into the frequency domain to generate input samples. Each category comprises 190 samples, with each sample set to a length of 1024. Subsequently, 70\% of these samples are randomly allocated as the training set, while the remaining samples are designated for the test set. Both the training and test sets are normalized by \textit{0-1} normalization as
\begin{equation}
        \textbf{x}_i =\left(  {\textbf{x}_i-\min(\textbf{x}_i)}\right ) / \left( {\max(\textbf{x}_i)-\min(\textbf{x}_i)}\right)
\end{equation}

Regarding data augmentation, three augmentations \cite{liVariationalAttentionBasedInterpretable2022} are applied to both the training and test sets to simulate the discrepancies encountered in real industrial environments:
\begin{equation}
        \begin{split}
                \textit{RandomAddNoise: } &\textbf{x}_i = \textbf{x}_i +  v \cdot 10 \cdot \mathcal{N} \left(0,\textrm{std}(\textbf{x}_i) \right)\\
                \textit{RandomScale: } &\textbf{x}_i = \mathcal{N} \left(1,v \right) \cdot \textbf{x}_i\\
                \textit{RandomMask: } &\textbf{x}_i = \textrm{mask}(\textbf{x}_i,d)\\
        \end{split}
\end{equation}
where $v$ and $p$ are hyperparameters of augmentation, and the \textit{mask} function set a segment with length $d$ of $\textbf{x}_i$ to $0$. The random probability of all three augmentations is set to 0.5.

Concerning model hyperparameters, the number of prototypes in PMN is set as $m=11$ according to the number of state classes, and the loss is configured as $(\lambda, \lambda_1,\lambda_2,\lambda_3)=(1,0.25,0.25,0.01)$ based on grid search. Training parameters include 50 epochs, a batch size of 128, Adam optimizer, and a learning rate of 0.001 with a 0.99 decay per epoch.

Besides the traditional evaluation on diagnostic accuracy, a novel dimensionless metric $R_\textrm{rps}$ is introduced to evaluate the representation learning ability. Denoting the mean vector of learned features belonging to $k$-th class as $\overline{{\textbf{z}}}_k$, the intra-distance $D_{\rm intra}$ could be calculated as
\begin{equation}
        \begin{split}
                D_{\rm intra} = \frac{1}{n} \sum\nolimits_{i=1}^{n} \sum\nolimits_{k=1}^{K} \left\|\textbf{z}_i-\overline{\textbf{z}}_k \right\|_2 \mathbb{I}(y_i=k)
        \end{split}
        \label{eq:intra}
\end{equation}
and the inter-distance $D_{\rm inter}$ could be calculated as

\begin{equation}
        \begin{split}
                D_{\rm inter} &= \frac{1}{K(K-1)} \sum\nolimits_{i=1}^{K} \sum\nolimits_{j=1}^{K} \left\| \overline{\textbf{z}}_i-\overline{\textbf{z}}_j \right\|_2.
        \end{split}
        \label{eq:inter}
\end{equation}
Hence, this novel representation metric $R_\textrm{rps}$ could be expressed as
\begin{equation}
        R_\textrm{rps} = D_{\rm intra} / D_{\rm inter}.
        \label{eq:rps}
\end{equation}
A smaller value $R_\textrm{rps}$ is more preferable, indicating this model has stronger capability in representation learning, i.e., smaller intra-class distances and larger inter-class distances.

        {\bfseries \itshape 2) Diagnostic analysis}: Seven additional models are employed for comparison with our proposed PMN. They are Transformer from literature \cite{liVariationalAttentionBasedInterpretable2022}, CNN, biLSTM, ResNet18, and DCAE from the benchmark \cite{zhaoDeepLearningAlgorithms2020}, as well as prototypical network~\cite{snellPrototypicalNetworksFewshot2017} and CNN-PML~\cite{meiCrossdomainOpensetFault2024} (CNN for feature extraction and PM-layer for classification).
The above seven models include prototype-matching based methods (i.e., prototypical network and CNN-PML), and three mainstreams used in current deep learning: convolutional-type (CNN, ResNet, DCAE), recurrent-type (BiLSTM), and attention-type (Transformer). Besides, DCAE is also the backbone autoencoder of PMN but use MLP as the classifier instead of the PM layer.

The diagnostic accuracies and representation metrics of the above eight models under four noise augmentation levels are presented in Table~\ref{tab:Acc-DCDS}. Generally, the proposed PMN exhibits competitive diagnostic performance and excellent representation learning ability compared to the other seven models.

\begin{table*}[t!]
        \centering
        \footnotesize
        \begin{threeparttable}
                \caption{Test Accuracy and Representation Metrics of Different Models on Conventional Fault Diagnosis Experiment. \label{tab:Acc-DCDS}}
                \begin{tabular*}{\hsize}{@{\extracolsep{\fill}}c|ccccc|ccccc}
                        \toprule[1pt]
                        \multirow{2}{*}{Model} & \multicolumn{5}{c|}{Accuracy (\%)} & \multicolumn{5}{c}{Representation   metric $R_\textrm{rps}$}                                                                                                                                         \\
                        & 0-0\tnote{\dag}                    & 0.1-100                                                      & 0.2-100        & 0.2-200        & Average        & 0-0            & 0.1-100        & 0.2-100        & 0.2-200        & Average        \\
                        \midrule[0.3pt]
                        CNN~\cite{zhaoDeepLearningAlgorithms2020}                    & 98.65          & 96.23          & 89.01          & 87.45          & 92.84          & 0.792          & 0.912          & 1.201          & 1.323          & 1.057          \\
                        BiLSTM~\cite{zhaoDeepLearningAlgorithms2020}                 & 99.26          & 98.11          & 92.66          & 90.36          & 95.10          & 0.631          & 0.863          & 1.256          & 1.365          & 1.028          \\
                        ResNet18~\cite{zhaoDeepLearningAlgorithms2020}                 & 99.63          & 97.68          & 92.74          & 92.83          & 95.72          & 0.361          & 0.457          & 0.594          & 0.645          & 0.514          \\
                        Transformer~\cite{liVariationalAttentionBasedInterpretable2022}            & \textbf{100.0} & 98.72          & 94.23          & 91.33          & 96.07          & 0.618          & 0.788          & 0.987          & 1.012          & 0.851          \\
                        DCAE~\cite{zhaoDeepLearningAlgorithms2020}                   & \textbf{100.0} & 98.89          & \textbf{95.63} & 94.63          & 97.29          & 0.271          & 0.329          & 0.416          & 0.474          & 0.372          \\
                        ProtypicalNet~\cite{snellPrototypicalNetworksFewshot2017}          & 99.50 & 98.31   & 94.66   & 93.45   & 96.48   & 0.465 & 0.631   & 0.895   & 0.981   & 0.743   \\
                        CNN-PML~\cite{meiCrossdomainOpensetFault2024}               & 99.52 & 98.76   & 94.57   & 93.60   & 96.61   & 0.511 & 0.577   & 0.752   & 0.793   & 0.658   \\
                        PMN (our method)                & 99.98          & \textbf{99.31} & 95.41          & \textbf{94.81} & \textbf{97.37} & \textbf{0.195} & \textbf{0.225} & \textbf{0.316} & \textbf{0.345} & \textbf{0.270}\\
                        \bottomrule[1pt]
                \end{tabular*}
                \smallskip
                \scriptsize
                \begin{tablenotes}
                        \item[\dag] This row represents different noise augmentations of $v$-$d$. and ``Average" represents the average result of all the conditions.
                \end{tablenotes}
        \end{threeparttable}
\end{table*}

Specifically, all eight models can achieve nearly 100\% accuracy under noise-free condition. However, as noise intensity increases, convolutional methods (CNN, ResNet), recurrent methods (BiLSTM), and attention-based method (Transformer) declines significantly, reaching approximately 91\% accuracy in \textit{0.2-200} noise augmentation. Based on prototype-matching, prototypical network and CNN-PML perform much better and their accuracy is near 93.5\% in \textit{0.2-200} noise augmentation. Conversely, the autoencoder structure of DCAE and PMN reinforces the integrity of information extraction through the encoding-decoding process, ensuring the robustness of feature extraction. As noise increases, their diagnostic accuracy exhibit comparatively less decline, decreasing to approximately 94.6\% in \textit{0.2-200} noise augmentation. Moreover, PMN, constructed by replacing the MLP in DCAE with a PM layer as the classifier, exhibits a marginally enhanced diagnostic accuracy compared to DCAE, demonstrating the effectiveness of the PM layer.

In terms of the representation metric $R_\textrm{rps}$, DCAE and PMN surpass the other six models owing to their autoencoder structure. Moreover, the proposed PMN achieves superior performance across all noise augmentation conditions, attaining the lowest average $R_\textrm{rps}$ value of 0.270. This indicates that PMN can extract more discriminative features than the other seven models, thereby facilitating the fault classification.

\begin{figure}[htbp]
        \centering
        \includegraphics[width=10 cm]{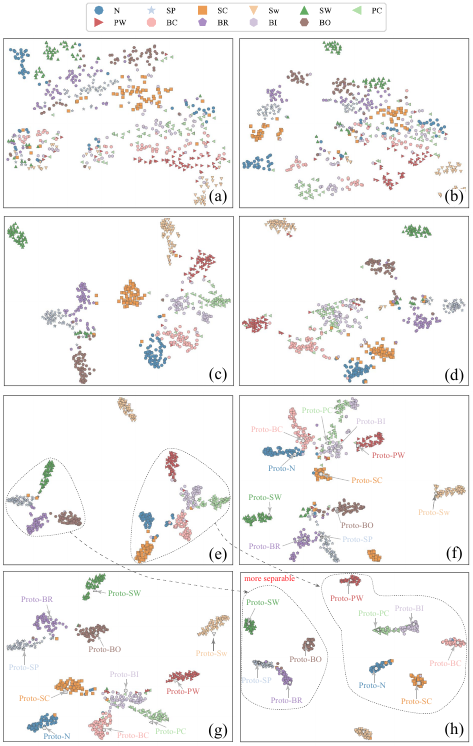}
        \caption{The t-SNE visualization of features extracted by different models on conventional fault diagnosis experiment with \textit{0.2-200} noise augmentation. (a) CNN. (b) BiLSTM. (c) ResNet18. (d) Transformer. (e) DCAE. (f) ProtypicalNet. (g) CNN-PML. (h) PMN.}
        \label{fig:t-SNE-DCDS}
\end{figure}

Although metric $R_\textrm{rps}$ succinctly reflects the capability of representation learning, the t-SNE visualization of the learned features is more evident. As illustrated in the Fig.~\ref{fig:t-SNE-DCDS}, the representations of CNN, BiLSTM, ResNet18, and Transformer are too indistinct to separate their categories, while DCAE performs significantly better, aligning with its lower $R_\textrm{rps}$ values in Table~\ref{tab:Acc-DCDS}. With prototype-matching mechanism, prototypical network and CNN-PML have relatively good t-SNE visualizations compared to the top four models, i.e., Fig.~\ref{fig:t-SNE-DCDS}(a)-(d). Due to the combination of  of autoencoder and prototype-matching, PMN takes a step further than all other models and its representations are tightly clustered around the prototype, making the learned features more separable. This enhancement is attributed to the PM layer, which explicitly conducts prototype-matching to further enhance the representation learning ability of the backbone autoencoder.

\begin{figure*}[htbp]
        \centering
        \includegraphics[width=17 cm]{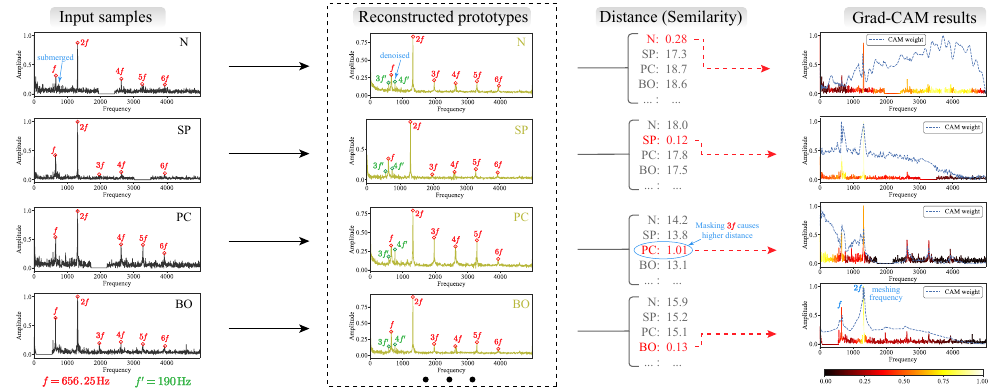}
        \caption{The interpretability analysis on conventional fault diagnosis experiment with \textit{0.2-200} noise augmentation. \textbf{Distance (3rd column)}: the first interpreting path explaining the clear classification logic of prototype-matching. \textbf{Reconstructed prototypes (2nd column)}: the second interpreting path explaining what typical fault signals look like. \textbf{Grad-CAM results (4th column)}: the third interpreting path explaining the attribution of high similarity through Grad-CAM. }
        \label{fig:Interpretability-DCDS}
\end{figure*}

{\bfseries \itshape 3) Interpretability analysis}: Utilizing the experimental schematic diagram in Fig.~\ref{fig:Exp1-rig} (b) and the electric motor speed of 1800 rpm, the meshing frequencies of fault-located planetary gearbox $f$ and subsequent healthy two-stage gearbox $f'$ could be calculated as $f=656.25 \; \textrm{Hz}$ and $f'=190 \;  \textrm{Hz}$, respectively. The interpretability analysis is demonstrated in Fig.~\ref{fig:Interpretability-DCDS}.

The classification logic compares the input sample with each prototype at the feature level and selects the fault class of the most similar prototype (lowest distance) as the prediction. Additionally, the \textit{mask} augmentation, blurring crucial frequencies, can introduce confusion for classification. For instance,  in the 3rd row with the PC fault, masking $3f$ frequency leads to the closest distance increasing from nearly 0.2 to 1.01.

The reconstructed prototypes depict what typical fault signals look like. Although the input samples are blured and the meshing frequency of subsequent two-stage gearbox $f'$ is submerged by noise augmentation, the reconstructed prototypes remain  resilient to noise and recover $3f'$ and $4f'$ from noise disturbance. This denoising and recovering characteristics of reconstructed prototypes underscore the promising application of PMN in noise senarios.

The Grad-CAM results reveal the key fault-related frequency responsible for the high similarity between the input signal sample and the matched prototype. Meshing frequency of planetary gearbox with its harmonic components, especially $f$ and $2f$, always get high contributions as shown in the 4th column. This indicates that the meshing frequency of planetary gearbox is the most crucial fault-related frequency for planetary gearbox fault diagnosis, which is consistent with the physical meaning of planetary gearbox fault.

\subsection{Domain Generalization Experiment of Bevel Gearbox}

\begin{figure}[htbp]
        \centering
        \includegraphics[width=10 cm]{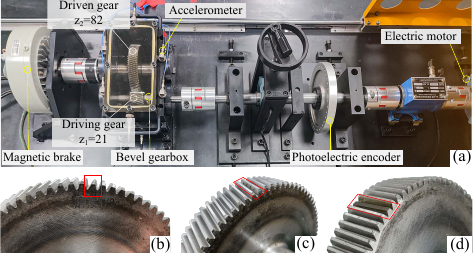}
        \caption{The experimental setup and fault types of the bevel gearbox. (a) The experimental setup. (b) W: Surface wear. (c) P: Pitting. (d) C: Tooth crack.}
        \label{fig:exp2-rig}
\end{figure}

{\bfseries \itshape 1) Data preparation}: The experimental setup and fault types are shown in Fig.~\ref{fig:exp2-rig}.
The accelerometer is positioned on the bearing end-shield at a sample frequency of 10 kHz. The vibration signals are collected under four conditions: $D_{L0}$, $D_{L1}$, $D_{H0}$ and $D_{H1}$, where subscripts $L$ and $H$ denote 1800 rpm and 2400 rpm, respectively. Subscripts $0$ and $1$ represent load and unload conditions. This dataset includes four states: normal (N), wear (W), pitting (P), and crack (C).

Vibration signals collected under various load and speed conditions exhibit distinct distributions. However, this discrepancy is attributed to operating conditions, and the underlying fault essence remains consistent. Consequently, treating each condition as a distinct domain, six cross-domain subtasks are formulated as Table~\ref{tab:Houde2-tansfer-setting} to assess the generalization ability of PMN. Set $m=4$, and other experimental settings are the same as the previous planetary gearbox experiment.

\begin{table}[htbp]
        \centering
        \footnotesize
        \begin{threeparttable}
                \caption{Subtask Settings of Domain Generalization Experiment. \label{tab:Houde2-tansfer-setting}}
                \setlength\tabcolsep{14pt} 
                \begin{tabular*}{\hsize}{@{\extracolsep{\fill}}cll}
                        \toprule[1pt]
                        Subtask name & Source domain & Target domain  \\
                        \midrule[0.3pt]
                        $T_1$     & $D_{L1}$ \& $D_{H0}$ \& $D_{H1}$         & $D_{L0}$          \\
                        $T_2$     & $D_{L0}$ \& $D_{H0}$ \& $D_{H1}$         & $D_{L1}$          \\
                        $T_3$     & $D_{L0}$ \& $D_{L1}$ \& $D_{H1}$         & $D_{H0}$          \\
                        $T_4$     & $D_{L0}$ \& $D_{L1}$ \& $D_{H0}$         & $D_{H1}$          \\
                        $T_5$     & $D_{L0}$ \& $D_{H0}$                  & $D_{L1}$ \& $D_{H1}$ \\
                        $T_6$     & $D_{L1}$ \& $D_{H1}$                  & $D_{L0}$ \& $D_{H0}$ \\
                        \bottomrule[1pt]
                \end{tabular*}
        \end{threeparttable}
\end{table}

{\bfseries \itshape 2) Diagnostic analysis}: The diagnostic results are presented in Table~\ref{tab:Acc-Houde2}. This domain generalization experiment presents higher diagnostic challenges, with no single method achieving overwhelming superiority. CNN performs best in $T_3$, while BiLSTM excels in $T_1$ and $T_6$. However, the proposed PMN has the optimal overall performance, achieving a highest average accuracy of 85.39\%.
In terms of $R_\textrm{rps}$, the proposed method outperforms all other models, showcasing overwhelming superiority across all tasks and achieving the lowest average value of 0.253.
In conclusion, the proposed method slightly outperforms other models in diagnostic accuracy and exhibits a significant advantage in representation learning.

\begin{table*}[htbp]
        \centering
        \footnotesize

        \caption{Test Accuracy and Representation Metrics of Different Models on Domain Generalization Diagnostic Experiment. \label{tab:Acc-Houde2}}
        \resizebox{\textwidth}{!}{
                \begin{tabular}{c|ccccccc|ccccccc}
                        \toprule[1pt]
                        \multicolumn{1}{c|}{}                                           & \multicolumn{7}{c|}{Accuracy (\%)} & \multicolumn{7}{c}{Representation metric $R_\textrm{rps}$}                                                                                                                                                                                                             \\
                        Model                                                           & $T_1$                              & $T_2$                                                      & $T_3$          & $T_4$          & $T_5$          & $T_6$          & Average        & $T_1$          & $T_2$          & $T_3$          & $T_4$          & $T_5$          & $T_6$          & Average        \\ \midrule[0.3pt]
                        CNN~\cite{zhaoDeepLearningAlgorithms2020}                       & 37.70                              & 94.93                                                      & \textbf{92.68} & 86.27          & 56.33          & 51.93          & 69.97          & 1.384          & 0.975          & 1.148          & 0.954          & 0.964          & 1.106          & 1.089          \\
                        BiLSTM~\cite{zhaoDeepLearningAlgorithms2020}                    & \textbf{86.62}                     & 97.44                                                      & 74.73          & 75.53          & 69.59          & \textbf{70.97} & 79.15          & 0.941          & 0.640          & 0.881          & 0.820          & 0.781          & 0.744          & 0.801          \\
                        ResNet18~\cite{zhaoDeepLearningAlgorithms2020}                  & 57.52                              & 80.63                                                      & 65.23          & 74.70          & 66.31          & 51.53          & 65.99          & 0.602          & 0.361          & 0.544          & 0.698          & 0.376          & 0.538          & 0.520          \\
                        Transformer~\cite{liVariationalAttentionBasedInterpretable2022} & 70.43                              & 80.46                                                      & 83.42          & 74.92          & 83.67          & 57.02          & 74.99          & 0.840          & 0.649          & 0.480          & 0.548          & 0.721          & 0.513          & 0.625          \\
                        DCAE~\cite{zhaoDeepLearningAlgorithms2020}                      & 80.85                              & 99.83                                                      & 75.05          & 76.73          & \textbf{96.02} & 61.51          & 81.66          & 0.538          & 0.319          & 0.518          & 0.239          & 0.215          & 0.571          & 0.400          \\
                        ProtypicalNet~\cite{snellPrototypicalNetworksFewshot2017}       & 53.98                              & 54.26                                                      & 79.78          & 68.56          & 53.91          & 60.60          & 61.85          & 0.619          & 0.469          & 0.573          & 0.556          & 0.514          & 0.560          & 0.548          \\
                        CNN-PML~\cite{meiCrossdomainOpensetFault2024}                   & 48.91                              & 52.77                                                      & 78.78          & 65.17          & 50.14          & 59.81          & 59.26          & 0.745          & 0.632          & 0.566          & 0.659          & 0.508          & 0.590          & 0.617          \\
                        PMN (our method)                                                & 75.35                              & \textbf{99.92}                                             & 75.00          & \textbf{99.45} & 95.45          & 67.19          & \textbf{85.39} & \textbf{0.281} & \textbf{0.119} & \textbf{0.396} & \textbf{0.131} & \textbf{0.123} & \textbf{0.470} & \textbf{0.253} \\
                        \bottomrule[1pt]
                \end{tabular}}
\end{table*}

\begin{figure}[htbp]
        \centering
        \includegraphics[width=10 cm]{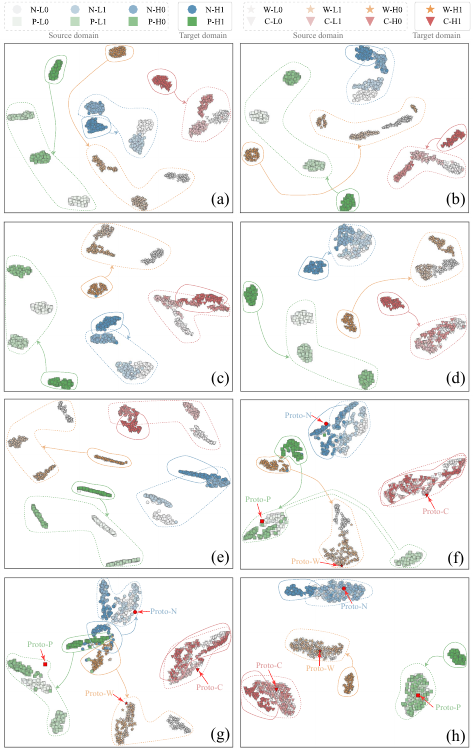}
        \caption{The t-SNE visualization of features extracted by different models under domain generalization task $T_4$. (a) CNN. (b) BiLSTM. (c) ResNet18. (d) Transformer. (e) DCAE. (f) ProtypicalNet. (g) CNN-PML. (h) PMN.}
        \label{fig:t-SNE-Houde2}
\end{figure}

The t-SNE visualization under $T_4$ is shown in Fig.~\ref{fig:t-SNE-Houde2}.
A good cross-domain representation should cluster the same fault type samples together, regardless of the domain discrapencies. However, other five models are greatly affected by domain discrepancies, with samples clustering distinctly based on faults and domains as Fig.~\ref{fig:t-SNE-Houde2}(a)-(e). Conversely, the proposed PMN, through explicit prototype-matching, actively encourages samples of the same class to cluster around the prototypes during the training process. Therefore, the representation of PMN effectively overcomes domain discrepancies, with samples of the same fault in different domains closely clustered together as Fig.~\ref{fig:t-SNE-Houde2}(h). Moreover, samples from target domain are also more proximate to the corresponding fault cluster of source domains.

\begin{figure*}[t!]
        \centering
        \includegraphics[width=17 cm]{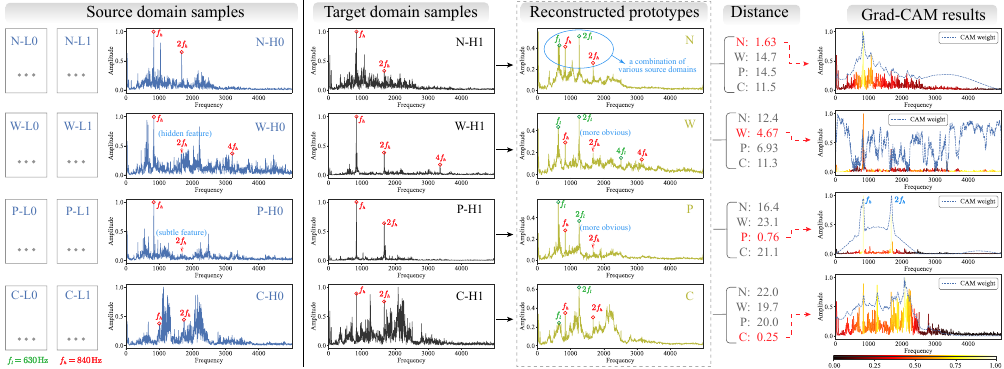}
        \caption{The interpretability analysis under domain generalization task $T_4$, includind three interpreting paths introduced in Fig.~\ref{fig:Interpretability-DCDS}.}
        \label{fig:Interpretability-Houde2Combine}
\end{figure*}

{\bfseries \itshape 3) Interpretability analysis}: Considering the tooth number of driving gear $z_1=21$ and the motor speeds, the meshing frequencies of low speed (1800 rpm) and high speed (2400 rpm) conditions could be calculated as $f_l= 630\; \textrm{Hz}$ and $f_h= 840\; \textrm{Hz}$, respectively.

The interpretability analysis is demonstrated in Fig.~\ref{fig:Interpretability-Houde2Combine}. The result is generally consistent with the conclusions drawn from the previous experiment. Firstly, the PM layer realizes fault prediction through prototype-matching as the classification logic. Secondly, the reconstructed prototypes demonstrates characteristics of denoising and uncovering weak signals. Lastly, the Grad-CAM results calculate the contribution of each frequency in the input sample to the high similarity, revealing the crucial role of meshing frequency and its harmonics ($f_h$ and $2f_h$ especially) in fault recognition.

Moreover, because $T_4$ involves three source domains ($D_{L0}$, $D_{L1}$, $D_{H0}$), the reconstructed prototype include both $f_l$ and $f_h$ and their harmonic components, as depicted in Fig.~\ref{fig:Interpretability-Houde2Combine} - 3rd column. It indicates that, from the model's perspective, the typical fault signal is a combination of various source domains when dealing with a multi-source domain scenario.

As an AI-driven interpretation of typical fault signals, the reconstructed prototypes demonstrate an effective extraction of crucial fault-related frequencies amidst noise or low energy, which could be regarded as a refection of broadening human understanding by  the interpretability research. Specifically, frequency $2f_h$ plays a crucial role for classification in the target domain, but in the source domains, they might be submerged (as \textit{W-H0} in Fig.~\ref{fig:Interpretability-Houde2Combine}) or extremely subtle (as \textit{P-H0} in Fig.~\ref{fig:Interpretability-Houde2Combine}). The PMN trained on the source domains adeptly captures this crucial frequency $2f_h$, making it more obvious in the reconstructed prototypes. This demonstrates that, gaining insights of ``\textit{what typical fault signals look like}'' from the model's perspective is highly meaningful and could enhance human understanding towards intricate mechanical fault classification. This not only improves the interpretability of AI but also serves as a feasible approach of AI-for-Science.

\section{Discussion}

\subsection{Discussion of Loss Coefficient and Distance Metric} \label{subsec:metric}

As indicated by Eq.~\ref{eq:Achitecture_PML}, the PM layer computes the distance between encoded feature and each prototype, where various distance metrics could be applied. Therefore, the effect of different distance metrics and loss coefficient is discussed here. The dataset is chosen as the bevel gearbox dataset used in the domain generalization experiment in the original paper. The experimental settings is consistent with the domain generalization experiment, comprising six sub-tasks as shown in Table~\ref{tab:Houde2-tansfer-setting}.

Notablely, the PM layer is equivalent to a linear model when using squared $L_2$ distance. The squared Euclidean distance metric $L_2$ used in PM layer could be denoted as
\begin{equation}
        d_{L_2}(f(\mathbf{x}),\mathbf{p}_k)=\parallel f(\mathbf{x})-\mathbf{p}_k\parallel _{2}^{2}
\end{equation}
which can be expanded as
\begin{equation}
\parallel f(\mathbf{x})-\mathbf{p}_k\parallel _{2}^{2}=f(\mathbf{x})^Tf(\mathbf{x})-2\mathbf{p}_{k}^{T}f(\mathbf{x})+\mathbf{p}_{k}^{T}\mathbf{p}_k.
\end{equation}
The first term  $f(\mathbf{x})^Tf(\mathbf{x})$ is a constant relative to the class and does not affect the softmax probabilities. The remaining terms can be expressed as a linear model:
\begin{equation}
-2\mathbf{p}_{k}^{T}f(\mathbf{x})+\mathbf{p}_{k}^{T}\mathbf{p}_k=\mathbf{w}_{k}^{T}f(\mathbf{x})+b_k
\end{equation}
where $\mathbf{w}_k=-2\mathbf{p}_k$ and $b_k=\mathbf{p}_{k}^{T}\mathbf{p}_k$.

\begin{figure}[htbp]
        \centering
        \includegraphics[width=10 cm]{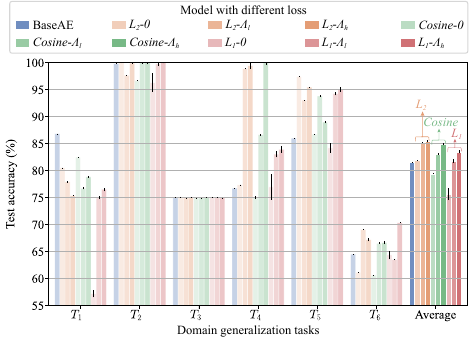}
        \caption{The diagnosis accuracy of different models with different losses under different Tasks. The first model ``BaseAE" is a simplified DCAE with only one FC layer as the classifier. The remaining nine models are represent PMN with three distance metrics ($L_2$, $Cosine$ and $L_1$) and three losses. $\lambda=0$ for all the models, and $0$ represents no extra loss, $\Lambda_l$  represents $(\lambda_1,\lambda_2,\lambda_3)=(0.05,0.05,0.002)$, $\Lambda_h$ represents $(\lambda_1,\lambda_2,\lambda_3)=(0.25,0.25,0.01)$.}
        \label{fig:Ana_Metric}
\end{figure}

The experimental results are presented as shown in Fig.~\ref{fig:Ana_Metric}. On average diagnostic accuracies, the metric $L_2$  demonstrates superiority compared to $L_1$  and $Cosine$, which outperforms other distance metrics under the same loss coefficients. When the loss coefficient is set to \textit{0}, the PMN-$L_2$ and the BaseAE have similar diagnostic accuracies. This effectively verifies the equivalence of the PM layer with distance metric $L_2$ to a linear model. Furthermore, with the increase of the loss coefficient, PMNs employing various distance metrics achieve higher accuracy, indicating the effectiveness of the proposed loss function, i.e., $R_1$, $R_2$ and $R_3$.

In summary, the PM layer with distance metric $L_2$ is equivalent to a linear model. However, the PM layer goes beyond this by actively promoting the mutual clustering of encoded features and their corresponding prototypes through the introduction of loss functions (i.e., $R_1$, $R_2$ and $R_3$), This leads to its competitive diagnostic performance and overwhelming representation learning capability.

\subsection{Discussion of The Number of Prototypes} \label{subsec:protonumber}

This classifier is distance-based in the low-dimentional feature-level space like \textit{mixture density estimation} (MDE) algorithm. The number of prototypes $m$ is equivalent to the number of density components in MDE, which is a key hyperparameter determined by the specific task. A natural question is whether it makes sense to use multiple prototypes per class instead of just one. Therefore, we design a multi-domain diagnostic task to explore the effect of the number of prototypes $m$.

The multi-domain diagnostic task means both the training and test datasets are sampled from an indentical distribution that containing multiple domains. The bevel gearbox dataset is selected to formulate this multi-domain diagnostic task as outlined in Table~\ref{tab:Houde2-multi-domain-setting}. This task comprises seven subtasks and four different numbers of domains.

\begin{table}[htbp]
        \centering
        \small
        \begin{threeparttable}
                \caption{Subtask Settings of Multi-Domain Diagnostic Experiment. \label{tab:Houde2-multi-domain-setting}}
                \begin{tabular*}{\hsize}{@{\extracolsep{\fill}}clc}
                        \toprule[1pt]
                        Task   name & Domain  information              & Domain number \\
                        \midrule[0.3pt] \vspace{1mm}
                        $Q_1$       & $L_0$                            & 1             \\
                        $Q_2$       & $H_0$                            & 1             \\
                        $Q_3$       & $L_0$ \& $H_0$                   & 2             \\
                        $Q_4$       & $L_1$ \& $H_1$                   & 2             \\
                        $Q_5$       & $L_0$ \& $H_0$ \& $L_1$          & 3             \\
                        $Q_6$       & $L_0$ \& $H_0$ \& $H_1$          & 3             \\
                        $Q_7$       & $L_0$ \& $L_1$ \& $H_0$ \& $H_1$ & 4             \\
                        \bottomrule[1pt]
                \end{tabular*}
        \end{threeparttable}
\end{table}

PMNs with four different numbers of prototypes, namely, \textit{4-8-12-16} (calculated as the product of the number of fault classes 4 and the number of domains \textit{1-2-3-4}), are employed for comparison. The models are denoted as \textit{PMN-x}, where \textit{x} represents the number of prototypes.

\begin{figure}[htbp]
        \centering
        \includegraphics[width=10 cm]{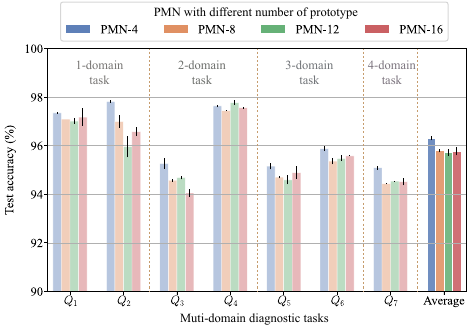}
        \caption{The accuracy of PMNs with four different numbers of prototypes on the multi-domain diagnostic task.}
        \label{fig:Ana_proto_M}
\end{figure}

The diagnostic results are shown in Fig.~\ref{fig:Ana_proto_M}. Overall, \textit{PMN-8}, \textit{PMN-12}, and \textit{PMN-16} exhibit similar diagnostic accuracy, but all lag noticeably behind \textit{PMN-4}. \textit{PMN-4} not only demonstrates clear superiority in the 1-domain ($Q_1$) subtask but also exhibits noticeable advantages in the 2-domain ($Q_3$) and 4-domain ($Q_7$) subtasks. The reason is that the proposed PMN is not a simple MDE but rather a combination of autoencoder and meta-learning. The autoencoder can nonlinearly map fault samples from different domains to the same fault cluster in the encoder process. Therefore, \textit{PMN-4} (with a single prototype per category, i.e.,  $m=K$) can achieve better performance without the need as MDE to adjust the number of density components based on the dataset characteristic.

\section{Conclusion}
To understand the neural network's classification logic and interpret ``\textit{what typical fault signals look like}'', we propose the prototype matching network (PMN), where prototype-matching is organically combined with autoencoder in IFD. The PMN not only explicitly applies human-inherent prototype-matching logic for classification but also effectively promotes the representation learning and depicts the typical fault signals from model's perspective. Additionally, Grad-CAM is adopted to explain the frequency sources contributing to the matched similarity. Conventional diagnosis and domain generalization experiments validate the competitive diagnostic performance and excellent representation learning ability of proposed PMN. Moreover, the learned typical fault signals demonstrate the capability to denoise and unearth subtle key features that are challenging for human experts to capture, providing a feasible way for AI-for-Science from interpretability research.

\normalem
\footnotesize
\bibliographystyle{elsarticle-num}
\bibliography{reference.bib}

\end{document}